\documentclass[preprint,1p,times]{elsarticle}

\usepackage{graphicx}
\usepackage{amssymb}
\usepackage{gensymb}
\usepackage{textcomp}

\usepackage{lineno}
\usepackage{xcolor}
\usepackage{hyperref}

\usepackage{pdflscape}

\usepackage{multirow}
\usepackage{booktabs}
\usepackage{colortbl}

\journal{Computer and Electronics in Agriculture}

\begin{document}

\begin{frontmatter}


\title{Using vis-NIRS and Machine Learning methods to diagnose sugarcane soil chemical properties}

\author[AGROSAVIA]{Diego A. Delgadillo-Duran}
\ead{ddelgadillo@agrosavia.co}
\author[AGROSAVIA]{Cesar A. Vargas-García}
\author[AGROSAVIA]{Viviana M. Varón-Ramírez}
\author[PUJ]{Francisco C. Calderón}
\author[AGROSAVIA]{Andrea C. Montenegro}
\author[AGROSAVIA]{Paula H. Reyes-Herrera}
\ead{phreyes@agrosavia.co}
\address[AGROSAVIA]{Corporación Colombiana de Investigación Agropecuaria, CI Tibaitatá, Bogotá, Colombia}
\address[PUJ]{School of Engineering, Pontificia Universidad Javeriana, Bogotá, Colombia }

\begin{abstract}

Knowing chemical soil properties might be determinant in crop management and total yield production. Traditional soil properties estimation approaches are time-consuming and require complex lab setups, refraining farmers from promptly taking steps towards optimal practices in their crops. Soil properties estimation from its spectral signals, vis-NIRS, emerged as a low-cost, non-invasive, and non-destructive alternative. Current approaches use mathematical and statistical techniques, avoiding machine learning frameworks. This proposal uses vis-NIRS in sugarcane soils and machine learning techniques such as three regression and six classification methods. 
The scope is to assess performance in predicting and inferring categories of common soil properties (pH, soil organic matter OM, Ca, Na, K, and Mg), evaluated by the most common metrics.  We use regression to estimate properties and classification to assess soil property status. In both cases, we achieved comparable performance on similar setups reported in the literature for property estimation for pH($R^2$=0.8, $\rho$=0.89), OM($R^2$=0.37, $\rho$=0.63), Ca($R^2$=0.54, $\rho$=0.74), Mg($R^2$=0.44, $\rho$=0.66) in the validation set.

\end{abstract}


\end{frontmatter}


\section{Introduction}
\label{S:1}

As the population grows, the demand for food continues to increase. However, unsustainable practices reduce the arable soil. Soils are dynamic systems that change in response to different natural and anthropogenic activities. Soil health must be a priority, particularly in agricultural practices, to increase productivity without affecting the soil. Soil chemical properties are important for agricultural production because they determine the amount of nutrients that plants uptake from soil to grow. It is essential to monitor soil quality through physicochemical analyses to provide a specific assessment looking towards sustainability \cite{bunemann2018soil}. 


Laboratory analysis of soils is widely used to know soil properties; it employs traditional chemical analyses that are expensive, time-consuming, and generates environmental contamination due to the number of chemical reagents used \cite{Nanni2006}. There is currently a growing demand to obtain immediate results. The search for alternatives for conventional laboratory analysis has allowed the NIRS technique to be a potential candidate. The usage of visible and near-infrared reflectance spectroscopy (vis-NIRS) of the electromagnetic spectrum emerges as a precision agriculture technique to monitor soil physicochemical characteristics at the field and laboratory level. This non-destructive analysis method is cost-effective,  provides rapid results, and can infer multiple components from a single spectrum. Also, it does not require chemical agents in the analysis procedure; thus, it is not harmful to the environment.

Vis-NIRS is a method based on the absorption of light by different materials in the near-infrared and visible region of the electromagnetic spectrum (400 - 2500 nm) \cite{Canasveras2012}\cite{viscarra2010spatial}. Materials absorb specific frequencies when irradiated with visible-NIR light. Absorption occurs when the incoming light frequency corresponds to the molecular vibration frequency of a constituent in the sample. A detector monitors the portion of the light reflected and decomposes into the components at different frequencies of the spectrum with the corresponding magnitudes.


Nevertheless, processing raw NIRS data requires (1) using advanced mathematical and statistical analysis to provide information on what and how much substance is present in the sample and (2) performing calibrations that guarantee accuracy. Soil Vis-NIR spectra are largely nonspecific because of the overlapping absorption of soil constituents. Complex absorption patterns generated from soil constituents and quartz need to be extracted from the spectra\cite{Stenberg2010}. One of the most used methods to estimate soil chemical properties from vis-NIRS is the Partial Least Squares Regression (PLSR) \cite{kawamura2017vis,stenberg2010visible,vibhute2018determination}, but it hides the nonlinear relationships between the spectrum and the soil constituents\cite{Yang2019}.

The usage of machine learning (ML) in soil science has increased in the last decade \cite{padarian2020machine}, also impacting the use of infrared spectral data to infer soil properties   \cite{ding2018machine}\cite{yang2019evaluation}. 
Recent studies use 140\cite{morellos2016machine}, 261\cite{nawar2019line}, and 523\cite{Yang2019} soil samples and adopt ML approaches (SVM, neural networks, random forest, and cubist)  to estimate organic carbon and matter, cation exchange capacity, pH, clay content, and nitrogen from fresh and processed samples vis-NIR. However, soil properties depend on soil-forming factors and processes in a specific region, and ML approaches performance depends on the training set. Therefore, models trained with data from a particular region are not easily extended in other locations.  These studies frequently focus on regression to estimate properties; however, a path towards classification is unexplored.

Colombia is a country with a prominent diversity of soils. In Colombia, IGAC institute have identified 11 of the 12 soil orders at 1:100.000 scale, according to USDA classification \cite{Suelos_y_tierras}. Previous studies in Colombia used NIRS, in an oxisol, to predict total carbon and total nitrogen and to incorporate these predictions for mapping using geostatistical techniques in a region of about 5100 hectares. \cite{camacho2014near}. Later, they found NIRS useful to predict also clay content in the same study area \cite{camacho2017near}. However, these studies use PLSR to estimate chemical properties \cite{camacho2017near}.


This work proposes to use ML regressors and classifiers to predict chemical properties from vis-NIRS soil data in sugar cane for panela soil samples. For the regression, we used and compared Linear Regression (LR), Support Vector Regression (SVR), and Least Absolute Shrinkage and Selection Operator (LASSO). We used an independent test set to validate the results and select the best model by using metrics like the correlation coefficient (CC), coefficient of determination ($R^2$) and mean squared error (MSE). Also, we used binary trees, linear and quadratic discriminant, Naive Bayes, Support Vector Machines (SVM), and k-Nearest Neighbors (kNN) algorithms for the classification. The class ranges are in agreement with soil fertility requirements. However, this definition led to classes with a few representative samples; thus, to compare classifiers, we use a cross-validation strategy and metrics such as the accuracy, Mathews correlation coefficient \cite{Chicco2020} and the confusion matrix.



This study uses vis-NIRS and ML approaches from sugarcane for panela Colombian soil samples (653 data points) with two specific objectives. First,  to evaluate the capacity of ML approaches to estimate six soil chemical properties: pH, organic matter (OM), calcium (Ca), magnesium (Mg), sodium (Na), and potassium (K) content.  We compare the selected ML model for each property with two scenarios that simulate traditional chemometric techniques (1) using the band with the highest regression coefficient(s) and (2) Partial Least Squares Regression (PLSR) \cite{Cozzolino2003}\cite{zornoza2008near}.  Second, to infer categories for soil properties to see whether this is a viable alternative.

\section{Materials and methods}
\label{S:2}

\subsection{Data}
\subsubsection{Study area and sample collection procedure}

This study uses  a data set derived from a previous study in the 
Hoya del río Suárez region in Colombia (Coordinates: 73\degree 22'\ - 73\degree 39'\ West longitude and 5\degree 53'\ - 6\degree 10'\ North latitude). This region covers an area of about 470 km\(^2\). Entisols, inceptisols, and vertisols characterize this region, according to the soil survey \cite{Suelos_y_tierras}. The area has two principals crops: sugar cane for panela agro-industry and grasslands.

\begin{figure}[htb!]
\begin{center}
\label{fig:study_area}
\includegraphics[width=0.6\columnwidth]{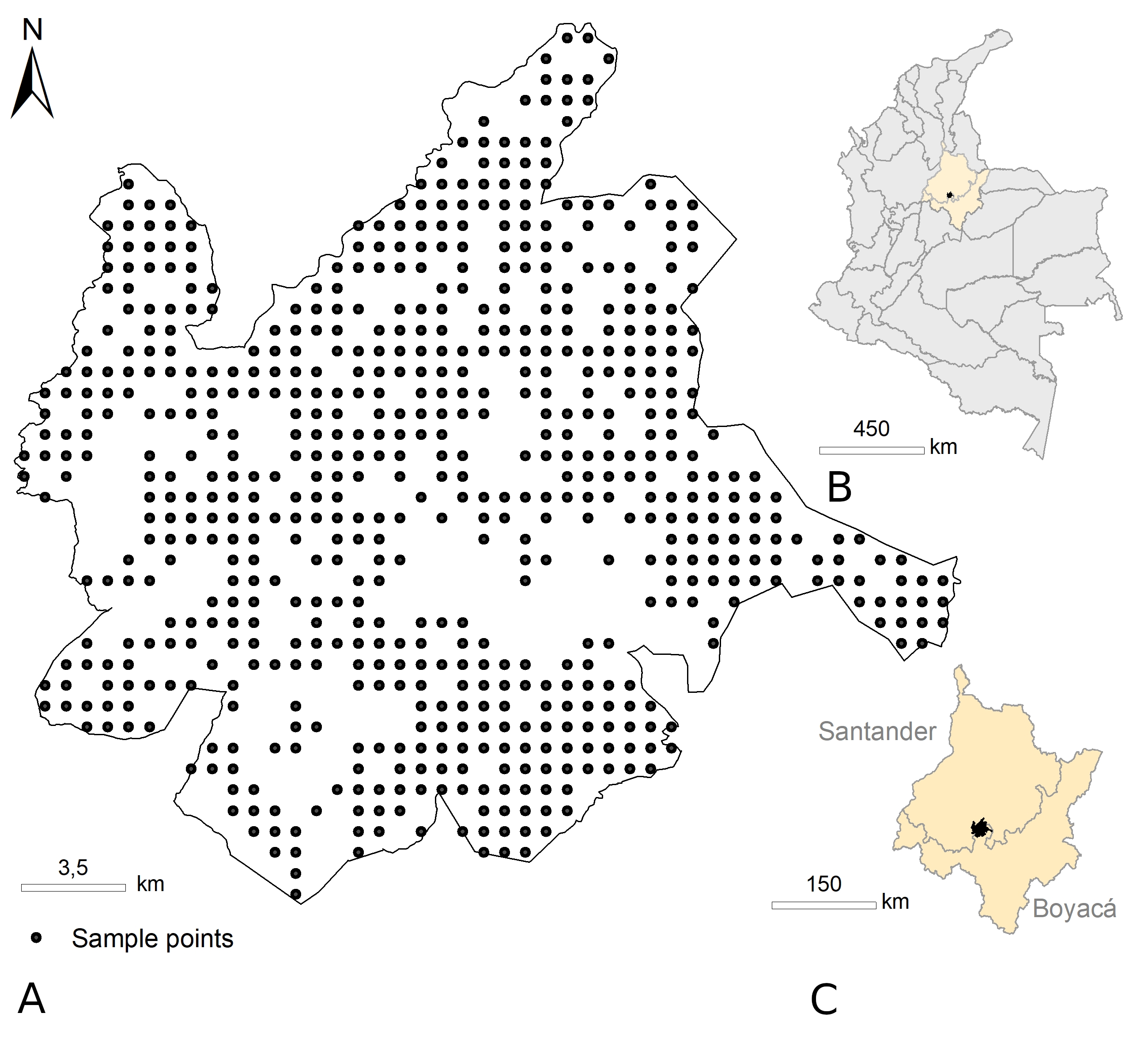}
\caption{\label{study_area} A. Sampling area for 653 points. Sample area in Santander and Boyaca district C. and in the country B.}
\end{center}
\end{figure}

The sampling stage occurred during 2015 and 2016; samples are from the surface to a depth of 20 cm. Each sample point represents four sub-samples collected and mixed. The sample design corresponds to a reticulate grid of 700 meters (Figure \ref{study_area}) wherein each point corresponds to four sub-samples to compose only one sample. This study uses a total of 653 samples.


\subsubsection{Chemical measurements} 

We dried and analyzed the samples for classic laboratory analysis. The laboratory procedure employed a pH meter with 1:2.5 soil-water suspension (NTC 5264, 2008), organic matter (OM) with Walkley and Black's wet digestion method. We used the ammonium acetate extraction method to measure exchangeable cations (Ca$^{+2}$, K$^+$, Mg$^{+2}$, and Na$^+$) by NTC 5349 - 2008 \cite{manualcane}.

\subsubsection{Vis-NIR spectroscopy}
The samples passed through a humidity homogenization process; samples were dried at 40°C (for 48-96 h, depending on the type of soil). Samples were placed in a 50 mm diameter annular cup and scanned from 850 to 2500 nm using a NIR spectrophotometer (FOOS-DS2500).

\subsection{Regression and classification strategy}
\begin{figure}[h!]
    \centering
    \includegraphics[width=10cm]{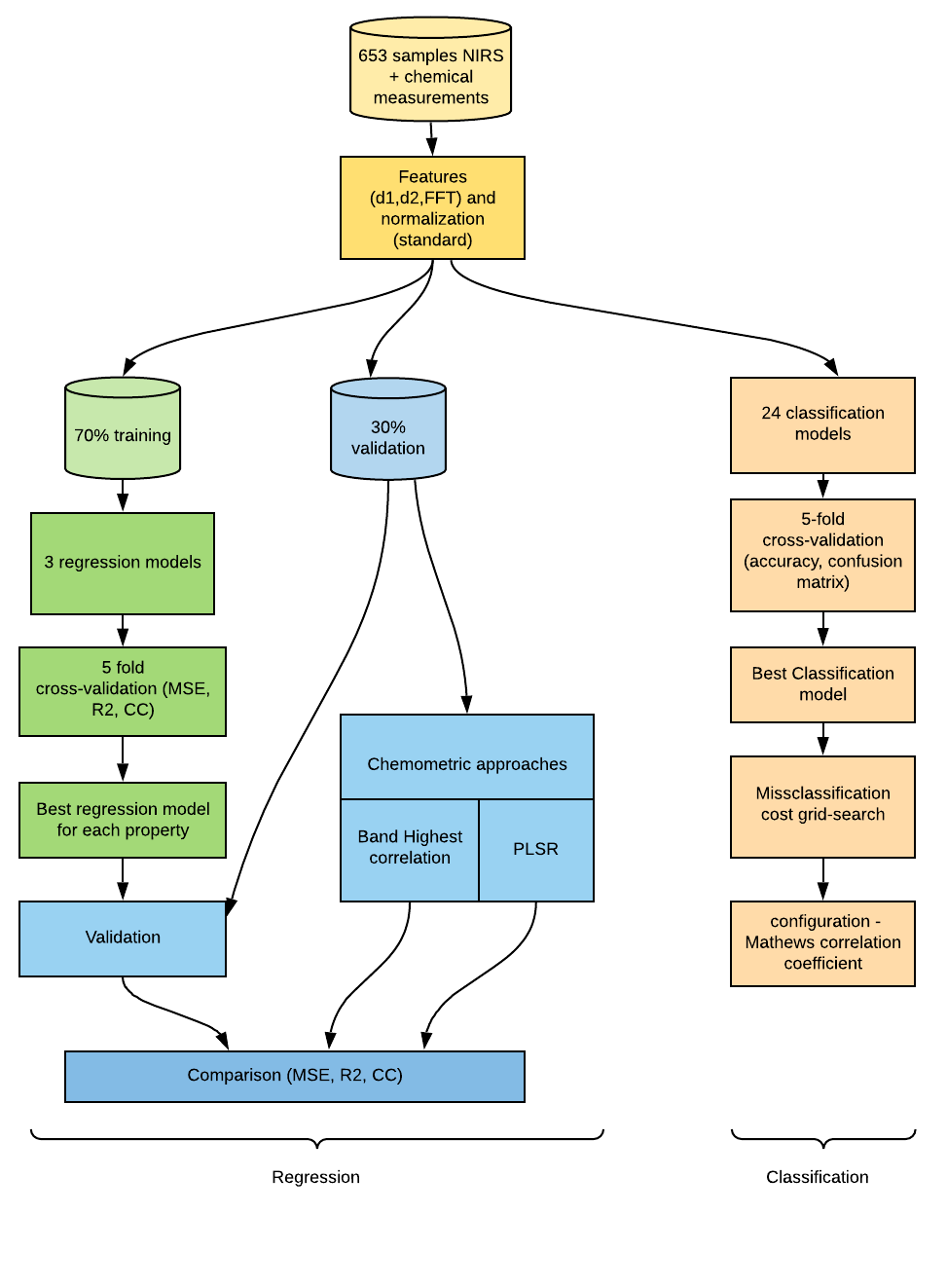}
    \caption{Strategy for regression and classification model selection from vis-NIRS data. Both start from the same data and features. At left, the steps that are taken to obtain the best regression model for each property;  and at right, the classification steps \label{strategy}}
    \label{fig:my_label}
\end{figure}

This study used the strategy presented in figure \ref{strategy} for regression and classification. First, the spectrum was transformed to obtain informative features (section \ref{feat}). Second, we followed two paths, one towards regression with three models from scikit-learn \cite{scikit-learn}in Python and another path towards classification with six classification methods from the Statistics and Machine learning toolbox in MATLAB. At follows, we describe the details for the ML regression (section \ref{reg}) and classification (section \ref{classifier}) models.
The coefficient of determination $R^2$ and correlation coefficient $\rho$ were used to compare and select the best regression model. We used the accuracy, confusion matrix, and Mathews correlation coefficient \cite{Chicco2020} to compare the classification models. 

\subsubsection{Features and  pre-processing \label{feat}} 
The Vis-NIR spectra for each sample cover the range between 400 and 2491 nm with steps of 8.5 nm (vector of 247 elements). We took each data point as a feature and applied transformations to the spectra, such as the first derivative (D1), second derivative (D2), and the Fast Fourier Transform (FFT). We applied standard normalization by feature in the whole samples dataset to ensure unit variance, and zero mean \cite{StdNorm} and concatenated each feature set, resulting in 247x4 = 988 features for every sample.

\subsubsection{ML regression \label{reg}}
The dataset contains 653 samples and 988 features for six soil properties. First, we randomly split this dataset 70\% (457 samples) for training (to evaluate and adjust parameters) and 30\% (196 samples) only for validation purposes.

\textbf{Regression models:} We evaluated three regression models from \cite{scikit-learn} such as (1) Linear Regression (LR) that fits a predictor based on a model of ordinary least squares linear regression \citet{scikit-learn}, (2) Support Vector Regression (SVR) with the linear kernel (based on libsvm library \citet{libsvm}, implements an Epsilon-Support Vector Regression predictor with free parameters $C=1$ and $\epsilon=0.1$), (3) LASSO by using cross-validation, implements a linear model that estimates sparse coefficients reducing the number of features, its fitting is iterative, and 5-fold cross-validation helps to select the best model (free parameters are $\epsilon=0.001$). 

\textbf{Cross-validation}: We selected the best model among the three previously mentioned using a \textit{5-fold} cross-validation in the 70 \% defined for training (457 samples) and the 988 features. We performed the selection by using the distribution, 5-fold (for the median and deviation), for the correlation ($\rho$), determination coefficients ($R^2$), and the mean squared error (MSE). However, for some properties, none of the regression ML appears promising ($\rho>$0.6); we did not proceed in this case because we consider that ML regression is not suitable for the property in this dataset. 

\textbf{Simulating chemometric approaches:} 
We simulate two chemometric approaches. For the first, we obtain the feature (from the 988) with the highest correlation coefficient to the target label; then, we used a model based on linear regression  with the best feature (LR-bf). For the second approach, we used Partial Least Squares Regression (PLSR), a method frequently used in the literature \cite{scikit-learn}. Although we used two up to eight components, the results reported correspond to the best tuned with six principal components and 10000 iterations.

\textbf{Comparison ML regression to simulated chemometric approaches}: We used the 30 \% test set (196 samples not used during the training) to compare the regression model selected against (1) the linear regression with the best feature (LR-bf), and (2) PLSR with six principal components.


\subsubsection{ML classifiers \label{classifier}}


\textbf{Classes}: We defined the target classes for the properties according to soil fertility requirements for sugarcane for panela: K (Low: $<$0.2, Medium: 0.2-0.4 and High: $>$0.4), Na (Acceptable: $<$1 and Not acceptable: $>$1), pH (acidity correction: $<$6, none correction: 6-7.3, alkalinity correction: $>$ 7.3), Mg (Low: $<$ 1.5, Medium: 3-5, High: $>$ 5), Ca (Low: $<$3, Medium: 3-6, High: $>$5), OM (Low: $<$3, Medium: 3-5, High $>$ 5). However, this definition conduced to imbalanced classes as shown in  Figure \ref{class} G.

\textbf{Classification Models}: 
We evaluated 24 configurations for ML classification models. These 24 classifiers can be arranged into six groups: (1) three based on binary trees (using a different number of splits, 4, 20, and 100), (2) Linear and Quadratic discriminant, (3) Kernel Naive Bayes with four kernels: Gaussian, Box, Epanechnikov, and Triangle, (4) Support Vector Machines or SVM with four different kernel configurations, linear, quadratic, cubic and gaussian, (5) K-Nearest Neighbors (KNN) with six different distance metrics (city-block, Chebyshev, Euclidean, Minkowski, Hamming, and Jaccard), and finally, (6) five ensemble-based architectures, Boosted trees, Bagged trees, Subspace discriminant,  Subspace KNN, and RUSBoosted Trees.

\textbf{Cross-validation}: Due to the class imbalance of the dataset, we opted to perform a \textit{5-fold} cross-validation in the entire dataset to select the best performing ML model.  The cross-validation gives us an estimate of the model accuracy trained and evaluated with all the data. This method generally results in a less biased and less optimistic evaluation of the model performance than other methods, such as train/test split. \cite{stuartjonathanrussell_2010_artificial}.

\textbf{Misclassification cost grid search}

Due to the class imbalance of the dataset, we perform a grid search over a penalty cost calculated from the confusion matrix to optimize the model hyperparameters. As usual, the confusion matrix represents how accurately the current model predicted the observation class when that particular observation was a part of the held-out fold in the cross-validation while the model is training. Therefore, the values of all the confusion matrices are integers that aggregate the complete dataset, as shown in Figures \ref{class} A-F.

This penalty cost was applied to all Type I and Type II errors in the Confusion Matrix. By default, the penalty is set to one in all errors and cero otherwise. Hence, all observations have the same penalty in the cost function. Changing this penalty affects the cost function of the model. 

We use a grid search to find the best-performing combination of penalty costs for all errors. In the models for pH, OM, Ca, Mg, K, we use a grid-search of 6 parameters corresponding to all type I and II errors on a three-class confusion matrix; each parameter varied between  $gs=\{1,2,...,7\}$  for a total of $117649$ different full cross-validation experiments. For Na, we optimize the two misclassification cost, we increased the grid search to $gs=\{1,2,...,150\}$ for a total of $22500$  cross-validations. Finally, we use the Mathews correlation coefficient or MCC as a metric to evaluate the performance due to the advantage of MCC over the accuracy or F1-score in unbalanced datasets \cite{Chicco2020}.

\subsubsection{Feature ranking}

Finally, we propose a feature ranking approach to unveil each wavelength or band's effects from the spectrum and the properties. We obtained and normalized scores for feature selection such as (1) the correlation coefficient, (2) LASSO ranking, (3) F-Score, and (4) variance. We only consider the derivatives for this procedure because the best bands for all properties belong to these transforms. The variance score, unsupervised, aids in filtering features with low variance and thus insufficient information. The LASSO ranking through regularization removes features with low information content and redundant, reducing dimensionality. Finally, F-Score and correlation coefficient rank the bands that have more effects on the target based on univariate regressions. We added these four scores and obtained a unique value to rank the features. We used the feature selection module form scikit learn \cite{scikit-learn} to obtain the scores.




\section{Results}


\subsection{Regression}
\begin{figure}[h!]
\begin{center}
\includegraphics[width=11cm]{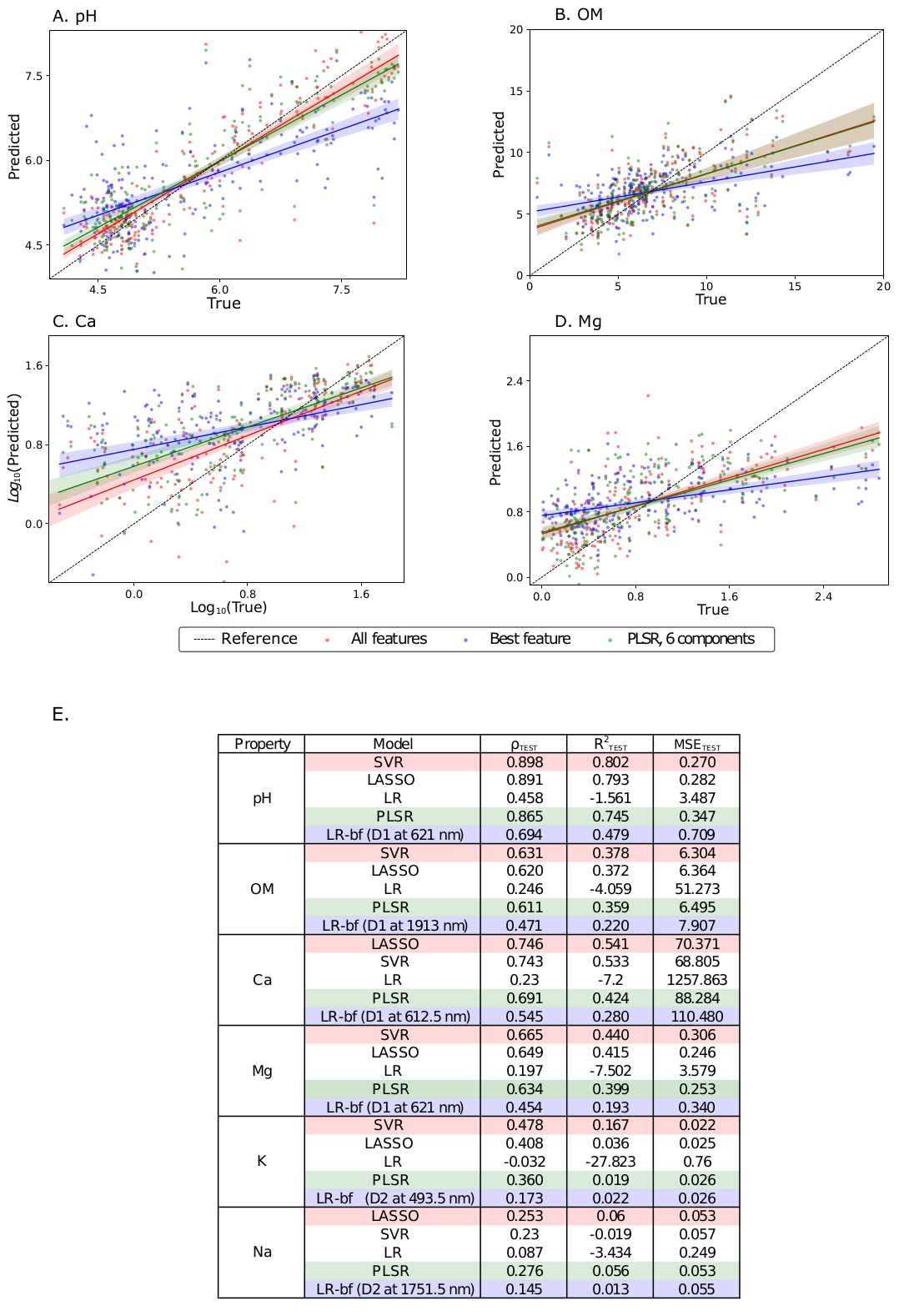}
\caption{\label{performance} \textbf{Regression results}: \textbf{A.} pH, \textbf{B.} Organic matter, \textbf{C.} Ca in log scale, \textbf{D.} Mg. For each property, we present the result with the best ML model (red). And the results simulate chemometric techniques such as the regression result with the band (blue) with the highest correlation and the PLSR (green). E. \textbf{Table with models comparison for property}: correlation $\rho_{test}$ and determination $R_{test}^2$ coefficients, and MSE in the test set (30\%). The models presented are the three result of the ML regression (background in red for the model with the best results), and the two approaches that simulate chemometric techniques (1) linear regression with the highest correlated band (background in blue) and (2) PLSR (background in green). Colors in the background correspond to the colors in Figures A-D}
\end{center}
\end{figure}

The best pH estimates result using an SVR regressor in the test set with a correlation between true and predicted $\rho=0.898$ ($R^2=0.802$). When using the feature that best correlates with pH, we get $\rho=0.694$ ($R^2=0.479$) using linear regression. LASSO performed slightly better than PLSR ($\rho=0.865$ and $R^2=0.745$). LASSO and PLSR regressor significantly improved the accuracy of the pH estimates, shown by the non-overlapping $95\%$ confidence interval of all three LASSO, LR-bf, and PLSR (Figure \ref{performance}A).

We got OM estimates correlated $\rho=0.62$ with ground true values ($R^2=0.37$) using the LASSO regressor. Comparing to the LR-bf regressor ($\rho=0.47$, $R^2=0.22$), LASSO and PLSR improved significantly the estimates (non-overlapping 95$\%$ confidence intervals) (Figure \ref{performance}.B). In the case of Ca, estimates using LASSO correlated $\rho=0.75$ with true values ($R^2=0.54$), showing a significant increase in accuracy if compared with LR-bf. LASSO also slightly improved PLSR estimates, although not significantly (Figure \ref{performance}.C).Also, Mg estimates using LASSO and PLSR were similar ($\rho=0.649$, $R^2=0.415$) and are significantly different LR-bf (Figure \ref{performance}.D).

Moreover, we tested several regression models on the remaining soil properties (K and Na), obtaining a correlation $\rho$ below $0.5$. Figure \ref{performance} E.  summarises regressor results and presents the best ML regressor and the results simulating chemometric approaches. We use linear regression to compare to a simple approach, but this method leads to the worst results in all the cases.

\label{S:3}

\subsection{Classification}

\begin{figure}[h!]
\begin{center}
\includegraphics[width=13.5cm]{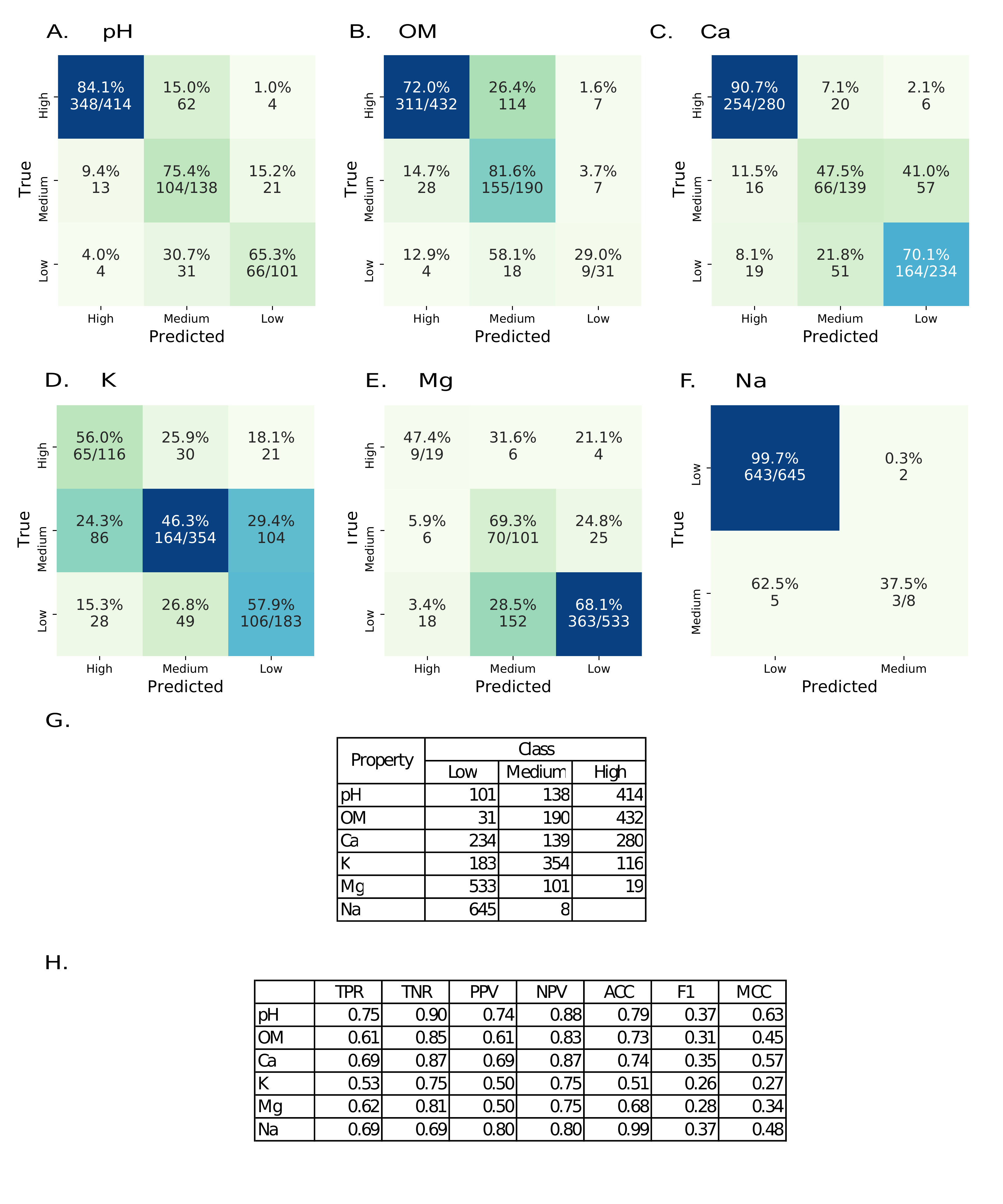}
\end{center}
\caption{\label{class} \textbf{Classification results}: Confusion matrix for each property \textbf{A.} pH, \textbf{B.} OM, \textbf{C.} Ca, \textbf{D.} K, \textbf{E.} Mg, \textbf{F.} Na. \textbf{G.} Table with number of samples divided by classes for each property, \textbf{H.} Table with metrics for the best classifier for each property. The reported metrics are True Positive Rate (TPR), True Negative Rate (TNR), Positive Predictive Value (PPV), Negative Predictive Value (NPV), Accuracy (ACC), F1 Score Harmonic mean of precision and recall (F1) and, Mathews Correlation Coefficient (MCC).}
\end{figure}

Remarkably SVM with a linear kernel was the best performing classifier of the 24 evaluated for all soil properties. The confusion matrixes of the best ML classifiers are shown in Figure \ref{class} A to F. As stated by Figure \ref{class} H, we obtain accuracies greater than 73$\%$ for pH, OM and Ca, and 68$\%$ for Mg. However, for OM, K, and Na some of the classes were under-represented by less than 5$\%$ of the total, this makes the precision results misleading.  i.e., for Na, a precision of 99$\%$ is not as good an indicator as appeared, since the medium class only has eight samples against 645 of the Low class. In this case, the F1 score, the True Positive Ratio TPR, and the MCC are preferred to compare the performance.


\subsection{Feature ranking}

\begin{figure}[hbt!]
\label{fig:feat_rank}
\centering
\label{fig:Features}
\includegraphics[width=1\columnwidth]{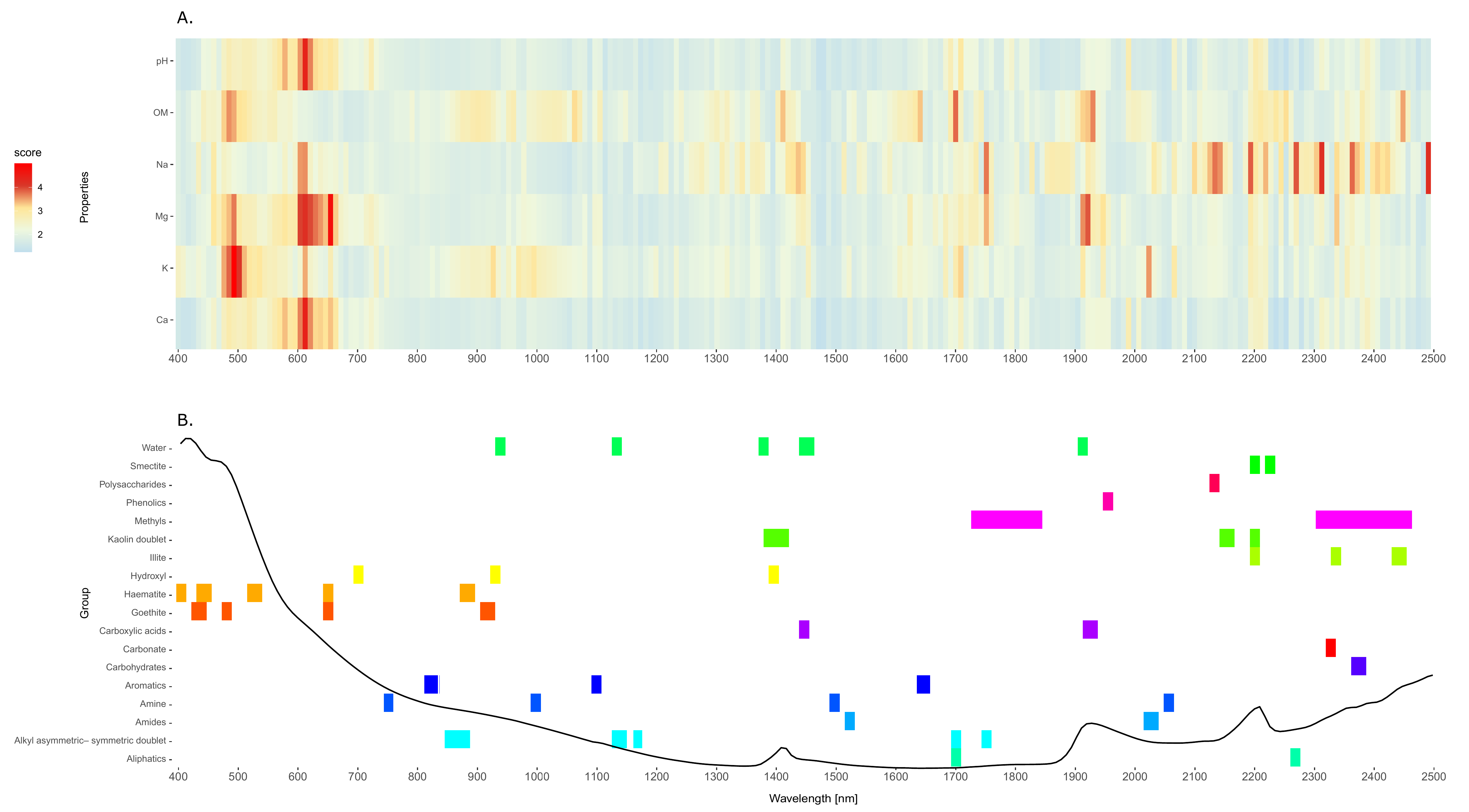}
\caption{\label{Features} \textbf{A.} bands with the highest correlation for each property \textbf{B.} NIRS spectra and bands with relative peak positions for soil costituents absorption}
\end{figure}

At last, Figure \ref{Features}.A shows the feature ranking for each property and region with distinct absorption (towards red). The visible 450-670nm range contains highly ranked features for all properties. pH and Ca have similar feature ranking heatmaps with the highest bands ranked around 600nm . Mg has a highly correlated range of 600-670nm, while K has a highly ranked area near 500nm. Thus, the visible spectrum bands are hihgly correlated to these properties.  Instead, Na presents highly ranked features between 2100-2400nm.

\section{Discussion}
\label{S:4}

Regression and vis-NIRS provide an alternative to soil property estimation, with an uncertainty higher than traditional laboratory analyses but with certain advantages (cost, timing, residues). The most frequently used method PLSR, has similar results to the best machine learning approaches but is slightly improved by them.

We proposed an alternative strategy for property estimation from soil samples by recasting the regression into a classification problem. We labeled conventional test results depending on the property to be estimated. We then implemented standard mappings of real property values into qualitative classes from literature. For our dataset, such mappings ended in unbalanced classes, with few samples (classes with less than $5\%$ of the samples). Surveyed ML classified the test samples mostly in the label with the largest training set. We marginally improved the classification of the under-represented labels by introducing weighted metrics in the cross-validation stage.We obtained the best results from a maximum margin method such as linear SVM, we believe that this result is due to the high tolerance to overfitting given by the cross-validation and the MCC. We also tested oversampling approaches to balance our training dataset synthetically. However, weighted metrics outperformed such an approach. These classifiers can be used as a qualitative assessment tool that might help optimal sampling design for further expensive conventional lab tests or initial intervention plans.


%

It is worth noting efforts in the literature attempt to use vis-NIRS data in proposals aiming towards soil diagnosis.
Viscarra et al. \cite{viscarra2010spatial} used Vis-NIRS, in sugarcane soils, to predict soil properties and moved towards a soil fertility index. They used 184 soil samples, PLSR, to estimate soil properties, in addition to  17 terrain attributes to derive the index. Awiti et al. \cite{awiti2008soil} used vis-NIR into an odds logistic model to classify soil into good, average and poor condition. The usage of vis-NIRS and ML is a rapid strategy that offers the possibility to diagnose soil conditions. This study is the first step to evaluate the performance of vis-NIRS, ML regressors, and classifiers, but we look forward to getting into soil diagnosis.


\textbf{Bands with the highest correlations and chemical hypotheses}

Regions for features highly ranked are centered near to 500, 600, 1400, 1700, 1900, 2200, and 2400 nm (Figure \ref{Features}). For the pH(H$_2$O), absorptions near 500 and 600 nm are primarily associated with some minerals containing hematite, and goethite \cite{morris1985, Stenberg2010}. At the same time, absorptions near 600 nm result from chromophores and the darkness of organic.                   
In the Vis-NIRS, the overtones and combination bands due to organic matter result from the stretching and bending of CO, CH, and NH groups \cite{BenDor1999}. The band around 1400 nm is linked to the vibration of OH and residual water in organic matter\cite{Reda2019}. On the other hand, the wavelengths at 1700 and 1930 nm are assigned to groups (C-H) and (C=O) that correspond to aromatic asymmetric alkyl-symmetric doublet and carboxylic acids, respectively \cite{Rossel2010}. These bands have been identified as essential bands for organic matter calibration \cite{Stenberg2010}. The band near 2200 nm can be attributed to metal–OH bend plus O–H stretch combinations of several clay minerals, among them illitic types\cite{Clark1990}, organic compounds, and carbonate. The wavelength at 2350 nm is related to Mg-OH \cite{Fang2018}.
Finally, in Figure \ref{Features} the region between 500 and 600nm has a high correlation with the chemical parameters analyzed. This could be related to both the dissolution mechanisms of iron oxides within the soils, and particularly within the rhizosphere (protonation, reduction, complexation) \cite{Schwertmann1991}; as well as the reactions that organic matter (humic acids and fulvic acids) with cations (Ca$^{+2}$, K$^+$, Mg$^{+2}$, and Na$^+$) (\cite{Ali, Sindelar2015,Wang,Yan2015,Droge2012}). Although there is no direct association between properties and the NIRS, highly classified characteristics could be associated with property components.

\section{Conclusions}
\label{S:5}

The combination of spectra, its first and second derivative, and ML regressors have the best accuracy results for pH, OM, Ca, and Mg soil content. Despite the estimation performance being close to reported in the literature, it is critical increasing the number of samples, adding soil samples with extreme values to enhance prediction power. PLSR has a comparable performance estimating chemical properties; although it has a similar performance to the best ML regression models, the best ML regressors outperform PLSR.

ML classifiers are a feasible strategy when ML regressors poorly perform. Also, ML classifiers can be used as a qualitative assessment tool for optimal sampling design.

The feature ranking approach enables the researcher to get insight into the bands that highly correlate with each property. It is essential to understand what is behind ML approaches; thus, feature ranking is the first step in getting back to the data.  


\section{Data availability upon acceptance}
The filtered datasets and scripts are archived at github (available upon acceptance).

\section{Acknowledegments}
Special thanks to Oscar Daniel Torres Rodríguez and Andrés Felipe Mariño Guerra for a preliminary study. We are also grateful for the project  \textit{243. Recomendaciones técnicas preliminares de manejo de suelos en ladera para el sistema de producción de caña panelera en la HRS} from AGROSAVIA and the Ministry of Agriculture and Rural Development (MADR) that obtained the data used in this study.%








\bibliographystyle{elsarticle-num-names}

\bibliography{main.bib}







\end{document}